\documentclass{article}
\usepackage{graphicx} % Required for inserting images
\usepackage{float}
\usepackage{amsmath}
\usepackage{hyperref}
\usepackage{indentfirst}
\usepackage{booktabs}
\usepackage{caption}
\usepackage{array}
\usepackage{subcaption}
\usepackage{geometry}
\usepackage{rotating}    % for sidewaystable environment
\usepackage{listings}
\usepackage{xcolor}
\usepackage{tikz}
\usepackage{tabularx}
\usepackage{doi}

\usetikzlibrary{
  positioning,
  shapes,
  arrows,    
  shadows.blur,
  fit
}
\usepackage{placeins}

\lstdefinestyle{mystyle}{
    backgroundcolor=\color{codegray},   
    commentstyle=\color{codegreen},
    keywordstyle=\color{codeblue},
    numberstyle=\tiny\color{codegray},
    stringstyle=\color{codered},
    basicstyle=\ttfamily\footnotesize,
    breakatwhitespace=false,         
    breaklines=true,                 
    captionpos=b,                    
    keepspaces=true,                 
    numbers=left,                    
    numbersep=5pt,                  
    showspaces=false,                
    showstringspaces=false,
    showtabs=false,                  
    tabsize=4
}

\title{Exploring Microstructural Dynamics in Cryptocurrency Limit Order Books:
Better Inputs Matter More Than Stacking Another Hidden Layer}

\author{%
  Haochuan (Kevin) Wang\\
  University of Chicago\\
  \texttt{haochuanwang@uchicago.edu}
}

\date{May 2025}

\begin{document}

\maketitle

\begin{abstract}
\noindent
Cryptocurrency price dynamics are driven largely by microstructural supply–demand imbalances in the limit order book (LOB), yet the highly noisy nature of LOB data complicates the signal extraction process. Prior research has demonstrated that deep‐learning architectures can yield promising predictive performance on pre‐processed equity and futures LOB data, but they often treat model complexity as an unqualified virtue. In this paper, we aim to examine whether adding extra hidden layers or parameters to “black‐box-ish” neural networks genuinely enhances short‐term price forecasting, or if gains are primarily attributable to data preprocessing and feature engineering. We benchmark a spectrum of models—from interpretable baselines, logistic regression, XGBoost to deep architectures (DeepLOB, Conv1D+LSTM)—on BTC/USDT LOB snapshots sampled at 100 ms to multi‐second intervals using publicly available Bybit data. We introduce two data‐filtering pipelines (Kalman, Savitzky–Golay) and evaluate both binary (up/down) and ternary (up/flat/down) labeling schemes. Our analysis compares models on out‐of‐sample accuracy, latency, and robustness to noise. Results reveal that, with data preprocessing and hyperparameter tuning, simpler models can match and even exceed the performance of more complex networks, offering faster inference and greater interpretability.
\end{abstract}

\section{Introduction}

Research on limit order books (LOBs)---the engines that match buy and sell orders in electronic markets---has provided deep insights into intraday liquidity provision, price impact, and market resilience, especially in equities, futures, and options (Hendershott et al., 2011). By exploring the dynamics of supply and demand, recognizing patterns behind order placements, LOB research has provided invaluable insights into the price discovery of assets, especially in high-frequency trading.

Cryptocurrency market diverges from traditional exchanges such as CME or Nasdaq. Centralized crypto exchanges like Bybit, Binance, and Deribit operate 24/7, have less regulatory oversight, and vast, globally distributed order flow. In this nonstop, high‑speed environment, systematic algorithmic strategies are essential for market participants to sustain vigilance and exploit fleeting opportunities. And one of the fundamental pieces in streamlining a strategy will be asset fair price prediction in a higher frequency domain. Millisecond-to-second price‑movement forecasts empower market makers to react and adjust theoretical quotes faster, isolating genuine liquidity imbalances from transient noise and other noises caused by flickering trades or order cancellations.

At the microstructural level, fundamental news drivers play a limited role, with price formation dominated by LOB dynamics. So, LOB has been the main source of data for trading traditional assets as well. Therefore, the result from exploring crypto spot pairs like BTC/USDT, which exhibit characteristic high turnover and deep liquidity, can be extended further to other liquid, volatile markets. Common preprocessing challenges—raw LOB cleaning, secondary feature engineering, flicker‐trade filtering, cancellation handling, and smoothing—are shared across asset classes. Moreover, unlike studies that rely on proprietary, third‐party‑filtered data, we work with raw historical snapshots from public crypto exchanges so the results of the research are more replicable and testable.

\subsection{Microstructural Focus}

We analyze 100\,ms LOB snapshots from Bybit, a leading CEX that offers limited access to historical data, omitting macro and cross-asset inputs. A typical LOB records multiple depth levels: bids represent quantities buyers will pay at or below specified prices; asks represent quantities sellers will accept at or above specified prices. Beyond the best bid/ask, successive levels reveal queued liquidity at less favorable prices. Plotting the first ten depths highlights where supply–demand asymmetries reside; cumulative depth imbalances often precede short‑horizon mid‑price moves.

\begin{figure}[H]
  \centering
  \begin{subfigure}[b]{0.48\linewidth}
    \centering
    \includegraphics[width=\linewidth]{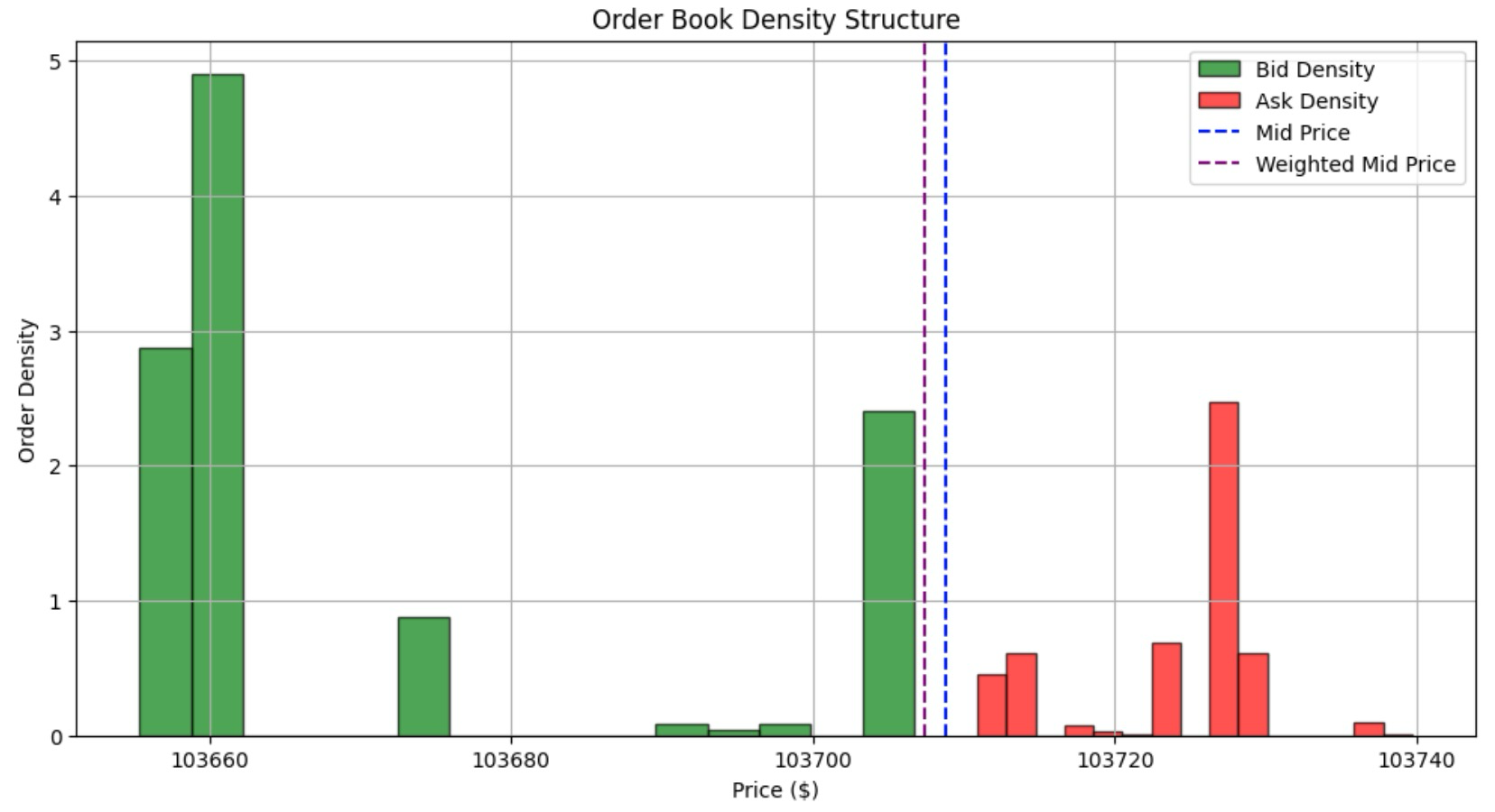}
    \caption{Raw LOB}
    \label{fig:raw-lob}
  \end{subfigure}
  \hfill
  \begin{subfigure}[b]{0.48\linewidth}
    \centering
    \includegraphics[width=\linewidth]{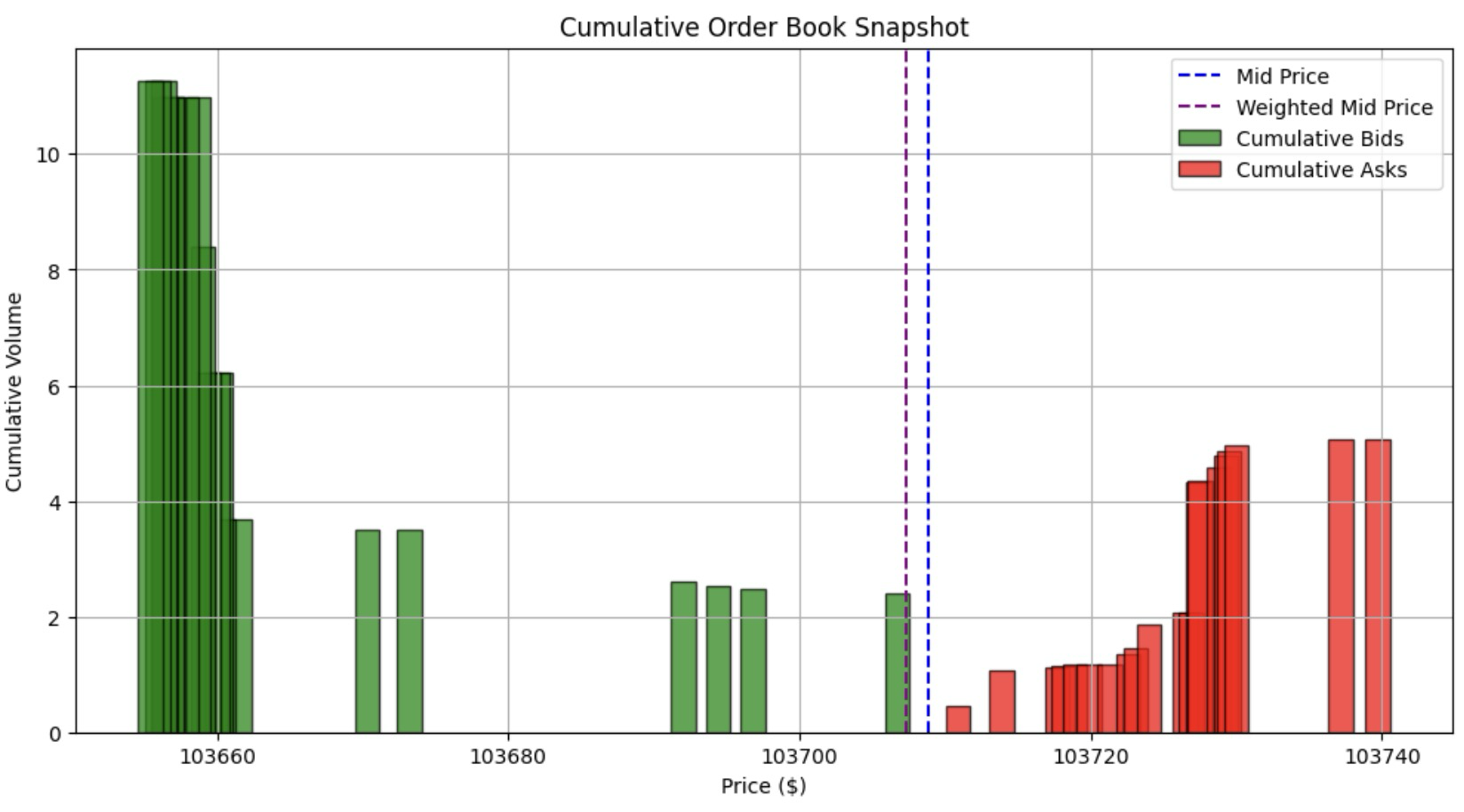}
    \caption{Cumulative LOB}
    \label{fig:cum-lob}
  \end{subfigure}
  \caption{Comparison of the raw and cumulative limit order books. Aggregating quantities across levels produces a cumulative order book, whose supply–demand asymmetries often precede short‑horizon mid‑price moves.}
  \label{fig:lob-comparison}
\end{figure}

And traditional in predicting price changes, we begin by constructing a set of economically intuitive, hand‑crafted features from raw LOB snapshots for example:
\begin{align}
m_{t-1} &= \frac{a^1_{t-1} + b^1_{t-1}}{2}
\quad\text{(Previous Mid-Price)} \\[0.5em]
I^1_t &= \frac{Q^b_{t,1} - Q^a_{t,1}}{Q^b_{t,1} + Q^a_{t,1}}
\quad\text{(First-Level Order Imbalance)} \\[0.5em]
I^5_t &= \frac{\sum_{i=1}^5 Q^b_{t,i} - \sum_{i=1}^5 Q^a_{t,i}}{\sum_{i=1}^5 Q^b_{t,i} + \sum_{i=1}^5 Q^a_{t,i}}
\quad\text{(Five-Level Aggregate Imbalance)} \\[0.5em]
\Delta m^w_t &= \sum_{i=1}^3 w_i \bigl(m^i_t - m^i_{t-1}\bigr)
\quad\text{(Weighted Mid-Price Change)}
\end{align}

where \(a^i_t\) and \(b^i_t\) are the level‑\(i\) ask and bid prices, \(Q^a_{t,i}\) and \(Q^b_{t,i}\) are the corresponding sizes, and weights \(w_i\) (e.g.\ \(w_i\propto1/i\)) sum to one. These features then will feed into baseline models (like logistic regression, XGBoost), which are trained and updated dynamically to capture a basket of most predictive signals.

Convolutional neural networks, by contrast, learn feature combinations automatically via stacked convolution and activation layers, reducing the need for manual engineering. To evaluate the trade‑off between model with simpler structure \(\&\) high interpretability with more complex structure, we compare six models—logistic regression, XGBoost, CatBoost, CNN+LSTM, a CNN+XGBoost hybrid, and DeepLOB—on both binary (up/down) and ternary (up/flat/down) tasks at horizons of 100ms, 500ms, and 1second. We further apply Kalman smoothing and Savitzky–Golay filtering to mitigate fleeting quotes and noise, and assess each model on out‑of‑sample accuracy and inference latency. This framework reveals whether architectural complexity delivers genuine forecasting gains or if careful preprocessing alone suffices.

\section{Literature Review}

\subsection{Logistic Regression Baselines}
Logistic regression is a standard approach for price prediction. Kercheval \& Zhang (2015) shows that with handcrafted features-like weighted kth-level mid prices and quantity imbalances—from LOB snapshots and applied linear models to classify the next mid-price movement. In this setup, each LOB state is vectorized and fed into a logistic classifier to predict upward, downward, or stationary movements.

Owing to its fast training and inference, logistic regression yields modest accuracy (48–64 \% on the FI-2010 dataset, depending on horizon) while offering low latency—crucial for reducing slippage in high-frequency trading. Its coefficients also lend interpretability by highlighting which features (e.g., spread, order imbalance) drive predictions. Although it cannot capture complex nonlinear patterns, many high-frequency signals exhibit approximately linear effects; this helps the model ignore nonlinear noise and remain more robust to overfitting on limited data compared to deeper networks (Ntakaris 2018).

\subsection{XGBoost and Gradient Boosted Trees}
Gradient boosted decision trees (GBDTs), such as XGBoost, offer a nonlinear alternative to linear models by fitting an ensemble of regression trees through gradient-based loss minimization. In practice, the same baseline inputs used by linear models, such as the top k bid and ask price and volumes levels, bid \(\&\) ask spread, and recent mid price returns, are augmented with engineered metrics like rolling, window volatility, order flow imbalance across multiple depth levels, and principal component summaries of the LOB snapshot. 

For instance, Malchevskiy (2019) processed about 4 million event records over ten trading days for five Nasdaq-100 tickers (AAPL, AMZN, GOOGL, INTC, MSFT), computing mid-price log-returns and top-10 level volume imbalances. Training CatBoost (a GBDT variant with ordered boosting) yielded at best of 63 \% accuracy on short-term movement prediction, depending on feature informativity. However, by relying on event data (completed transactions) rather than uniform snapshots, this approach bypasses cancellations and flickering orders—real-market noise—but introduces irregular time gaps (e.g., 100 ms vs. 1000 ms) that hinder deployment in systematic strategies expecting consistent LOB inputs.

\subsection{CNN+LSTM Deep Learning Models}
Deep learning approaches combine convolution neural networks (CNNs) with long short-term memory (LSTM) layers to automatically extract spatial and temporal features from LOB snapshots. Instead of relying on handcrafted metrics, a CNN applies one-dimensional filters across each snapshot—learning, for example, weighted imbalances across multiple levels—while an LSTM captures how these patterns over time. Tsantekidis et al.\ (2018) pioneered this architecture (DeepLOB), demonstrating that learned features often outperform traditional inputs. However, these automatically derived representations lack interpretability, making it difficult to understand which economic signals drive predictions and increasing the risk of overfitting on spurious patterns.

\subsection{Classification Framing (Binary vs. Ternary)}
Previous studies typically cast mid‑price movement forecasting as either a binary (up/down) or a ternary (up/flat/down) classification problem. Under the Efficient Market Theorem, all observable order‑book information is instantaneously incorporated into asset prices, producing the empirically observed 50:50 split in high‑frequency mid‑price moves, and people will theoretically not be able to make profits, as the market has already incorporated past information into pricing; thus, we should see only 50 \(\%\) prediction accuracy, equivalent to random guessing.

However, although a binary split is intuitive, some researchers argue that ignoring a “stationary” class can cause the model to misclassify minor fluctuations as larger up or down movements. Ternary classification therefore introduces a stationary range around zero returns (e.g., $-\varepsilon \le \Delta p / p \le +\varepsilon$). Although this complicates defining class boundaries, it reduces false positives, which is crucial for low‑latency strategies (Zhang et al., 2019). Choosing $\varepsilon$ requires balance: if $\varepsilon$ is too large, one class dominates (e.g., 95\% stationary), yielding a trivial “always predict stationary” model with 95\% accuracy; if $\varepsilon$ is too small, noise from fleeting quotes pollutes training. Inspired y Sirignano \& Cont (2019), in our project we will tune $\varepsilon$ to achieve roughly equal class frequencies when we do ternary split and apply inverse‑frequency weighting in the loss so that rare “up” and “down” events incur proportionally greater penalty so we can reduce the bais cuased by class inbalance when feeding our data in supervised learning models and selected neural networks.

\subsection{Denoising \& Preprocessing}
The LOB’s low signal‐to‐noise ratio is intrinsic to snapshot data (Hendershott et al., 2011). Our predictive research aims to produce price‐trend forecasts that can be the foundation for an executable trading strategy. Within snapshot‐based LOB data—as opposed to trade‐only datasets—we often observe price moves caused by flickering orders and rapid submissions and cancellations by market makers, which carry little information and should not be learned by our model.

Besides, many academic studies use preprocessed vendor datasets or lack a detailed walkthrough of the preprocessing process (FI‑2010, B3). From their methodological summaries, they apply per‐day z‑score normalization (Tsantekidis et al., 2017) to stabilize heavy‑tailed volumes and day‑by‑day scaling (BMF, 2019) using previous‑day statistics to avoid lookahead bias.

We extend from this and apply two time series filters before feature scaling. The first type is the Kalman filter (Wu et al., 2011), which aims to smooth mid-price and depth‐level features under a simple local‐random‐walk+noise model. The second option is Savitzky–Golay filter (Savitzky \& Golay, 1964), which is a smoothing technique aiming to preserve local curvature while suppressing noise in the LOB data.

Besides the above-mentioned filtering techniques, there are details in doing the model that help create better training data before feeding it into our network. For instance, we use a balancing model in creating three state labeling classes, or having two classes, ups and downs, but carefully set the threshold so we still include the majority of the data. And due to the noisy nature, we will experiment with predicting mid-price changes in different time frames to smooth out fluctuations. For instance we will adjust the time resolution from 100\,ms to 1\,s, so we have be able to focus on actual movements instead of moves caused by flickering trades and some of the trivial dynamics in the market but also not extending the prediction period too long which cause each frequency’s orderbook carry less signals.

\section{Data Set}
All LOB data in this study were obtained from Bybit’s public historical data portal\footnote{\url{https://www.bybit.com/derivatives/en/history-data}}. We focus on a single trading day, \texttt{2025-01-30}, sampled at 100ms intervals. Each snapshot is a JSON record containing a precise millisecond timestamp and the top 200 bid and top 200 ask levels, each with price and quantity. Because these are raw LOB snapshots, some contain fewer than 40 or even 5 levels; in such cases, missing entries are represented as NaN placeholders.

\section{Methods}

\subsection{Breakdown of the selected models:}

\paragraph{CNN + CatBoost.}  We first train a 1D Convolutional Neural Network to extract lower-dimensional embeddings from the input LOB data. The CNN consists of two convolutional layers with ReLU activations and batch normalization, followed by a global max pooling layer and a dense projection to a 64-dimensional embedding space. Then, this gradient‑boosted decision‑tree ensemble model, Catboost, is trained on these embeddings with inverse‑frequency class weights to improve robustness against noise and imbalance.

\begin{figure}[H]
    \centering
    \includegraphics[width=0.7\linewidth]{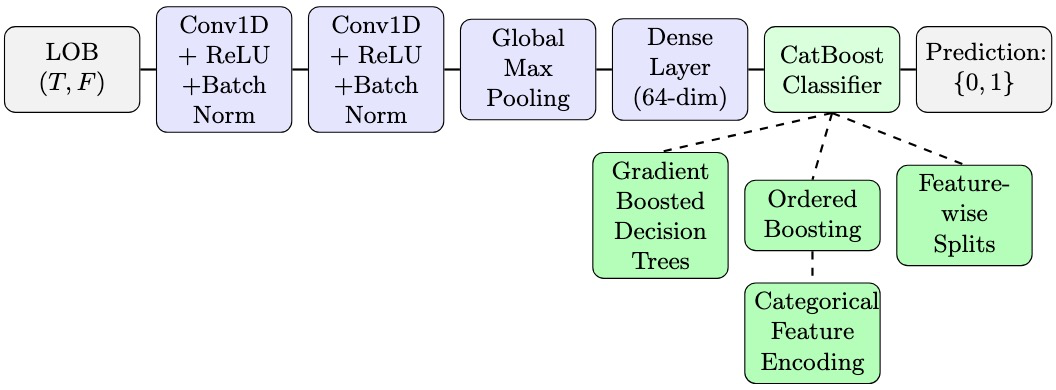}
    \caption{CNN + CatBoost: Pass embedding to CatBoost then performs ordered boosting}
    \label{fig:enter-label}
\end{figure}

\paragraph{DeepLOB.}  
DeepLOB reshapes each $(T \times F)$ input into a $(T \times F \times 1)$ tensor here T notate time and F notate layers of orderbook and applies three Conv2D blocks (kernel $1\times3$) with Batch Normalization and LeakyReLU activations to capture cross‑level interactions in price and volume. The output is flattened along spatial dimensions and fed into an LSTM(64) layer, which models sequential dependencies over the past 100 time steps. Trained with focal loss and class weights, this architecture is designed to focus on hard-to‑predict minority movements while retaining the ability to learn complex microstructural patterns.

\begin{figure}[H]
    \centering
    \includegraphics[width=0.7\linewidth]{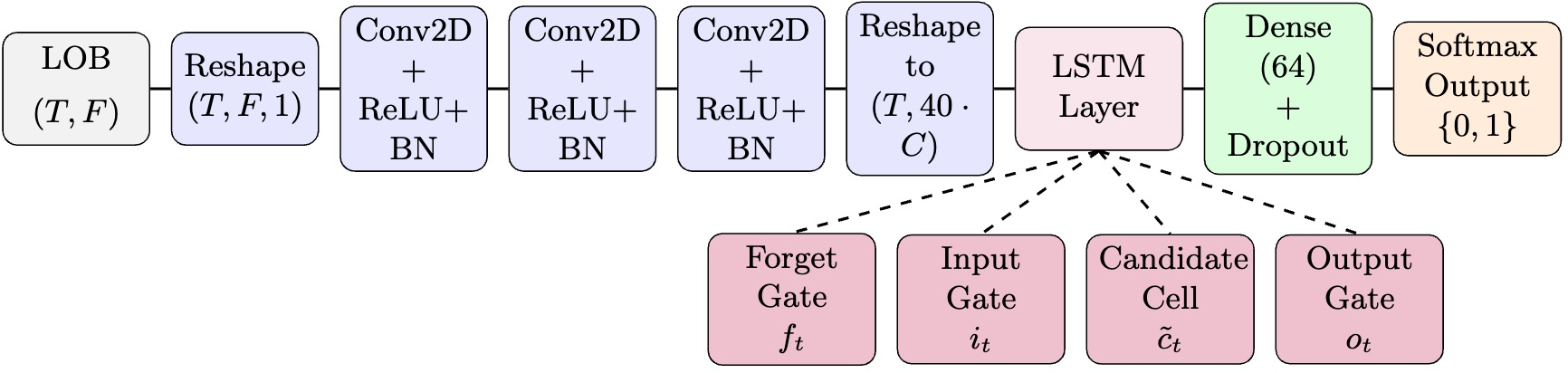}
    \caption{DeepLOB: Three CNN layers processed by an LSTM}
    \label{fig:enter-label}
\end{figure}

\paragraph{Simpler CNN + LSTM.}  
This streamlined architecture uses a single Conv2D layer, instead of three like DeepLOB, and then with Batch Normalization and SpatialDropout2D to regularize across price levels while extracting local imbalances. The resulting feature map is reshaped to $(T,\;\text{features})$ and processed by a bidirectional LSTM(32) to capture both forward and backward order‑flow autocorrelations.

\begin{figure}[H]
    \centering
    \includegraphics[width=0.7\linewidth]{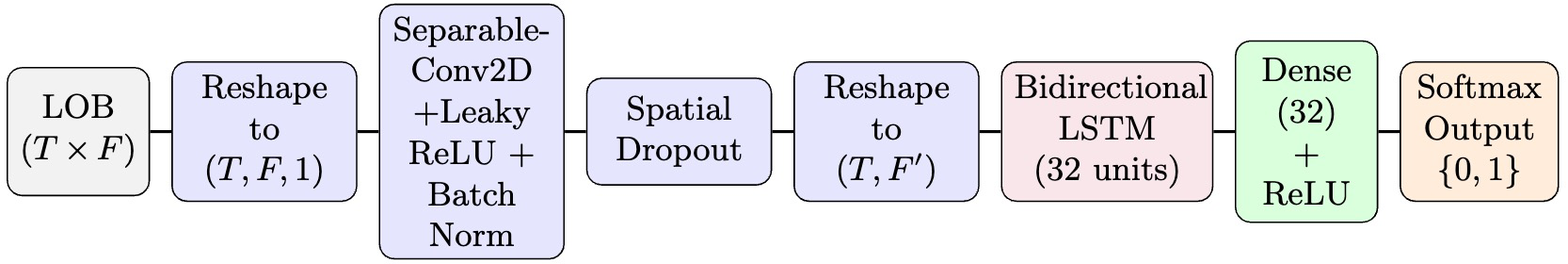}
    \caption{Simpler CNN + LSTM: One CNN layer processed by an LSTM}
    \label{fig:enter-label}
\end{figure}

\paragraph{XGBoost.}  
In the purely tree‑based approach, each $(T \times F)$ sequence is flattened into a single vector of length $T\cdot F$. An XGBClassifier with \texttt{multi:softprob} objective is trained on these vectors, using sample weights to correct class imbalance and early stopping on a held‑out validation set. Grid search over \texttt{n\_estimators} and \texttt{learning\_rate} tunes the bias–variance trade‑off. Despite its simplicity, this model excels at tabular LOB data and is a strong benchmark against more complex neural architectures.

% \begin{figure}[ht]
% \centering
% \begin{tikzpicture}[
%   font=\small,
%   input/.style={draw, rounded corners, fill=gray!10, minimum height=1.1cm, text width=1.6cm, align=center},
%   boost/.style={draw, rounded corners, fill=cyan!15, minimum height=1.1cm, text width=1.7cm, align=center},
%   intern/.style={draw, rounded corners, fill=cyan!30, minimum height=0.9cm, text width=2.6cm, align=center},
%   output/.style={draw, rounded corners, fill=orange!15, minimum height=1.1cm, text width=2.0cm, align=center},
%   arrow/.style={thick, -{Stealth[length=4pt]}},
%   node distance=0.45cm and 0.3cm
% ]

% % Main nodes
% \node[input] (input) {LOB \\$(T \times F)$};
% \node[boost, right=of input] (xgb) {XGBoost\\Classifier};
% \node[output, right=of xgb] (out) {Prediction:\\$\{0,1\}$};

% % Internals
% \node[intern, below=0.5cm of xgb, xshift=-3.5cm] (trees) {Gradient Boosted\\ Decision Trees};
% \node[intern, right=of trees] (depth) {Gird Search: Max Depth\\(e.g. 6)};
% \node[intern, right=of depth] (rate) {Gird Search: Learning Rate\\(e.g. 0.03, 0.1)};

% % Arrows
% \draw[arrow] (input) -- (xgb);
% \draw[arrow] (xgb) -- (out);

% % Internal arrows
% \draw[arrow, dashed] (xgb.south) -- (trees.north);
% \draw[arrow, dashed] (xgb.south) -- (depth.north);
% \draw[arrow, dashed] (xgb.south) -- (rate.north);

% \end{tikzpicture}
% \caption{XGBoost Only architecture: LOB input is passed through multiple rounds of gradient-boosted decision trees, controlled by learning rate, and depth}
% \label{fig:xgboost_pipeline}
% \end{figure}

\begin{figure}[H]
    \centering
    \includegraphics[width=0.45\linewidth]{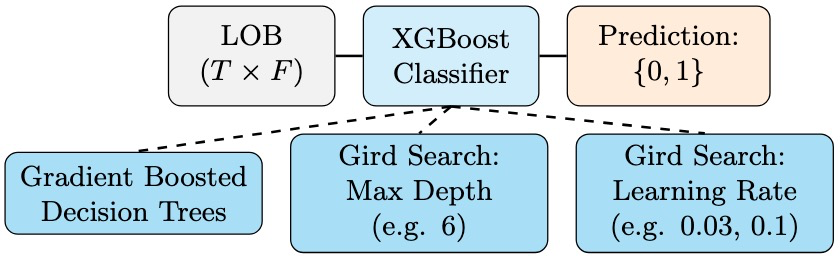}
    \caption{XGBoost Only architecture: LOB input is passed through multiple rounds of gradient-boosted decision trees, controlled by learning rate, and depth}
    \label{fig:enter-label}
\end{figure}

\paragraph{CNN + XGBoost.}  
Identical to the CNN in the CatBoost pipeline, this architecture trains two Conv1D layers with ReLU and Batch Normalization, followed by GlobalMaxPooling and a Dense embedding layer. Instead of CatBoost, an XGBoost classifier is applied to these embeddings.

\begin{figure}[H]
    \centering
    \includegraphics[width=0.7\linewidth]{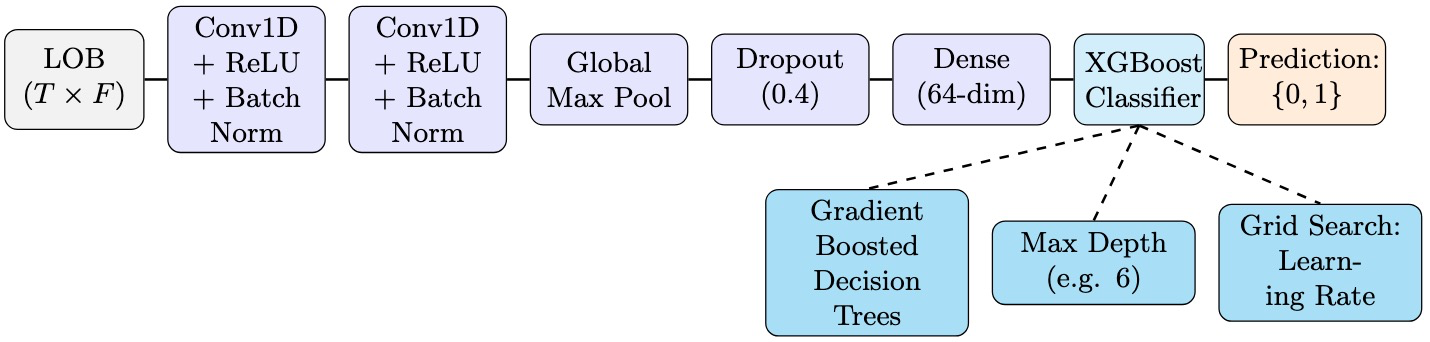}
    \caption{CNN + XGBoost architecture: a feature extractor network learns embeddings from limit order book data, which are passed to XGBoost with parameter tuning and early stopping.}
    \label{fig:enter-label}
\end{figure}

\subsection{Denoising via Savitzky–Golay and Kalman Filters}
Before feeding the data into the model architecture, we first preprocess and engineer the input features with the goal of denoising the raw limit order book (LOB) data. The LOB can be represented as a time series sampled at 100ms intervals, where each timestamp corresponds to a snapshot of the book up to depth k. These snapshots are inherently noisy due to the high-frequency nature of trading activity and microstructure effects. Rather than repeatedly training neural networks on such noisy data, we explore the effectiveness of two denoising techniques—Savitzky–Golay smoothing and Kalman filtering—in improving the model’s ability to extract meaningful signals from the noise.

\begin{figure}[H]
    \centering
    \includegraphics[width=0.6\linewidth]{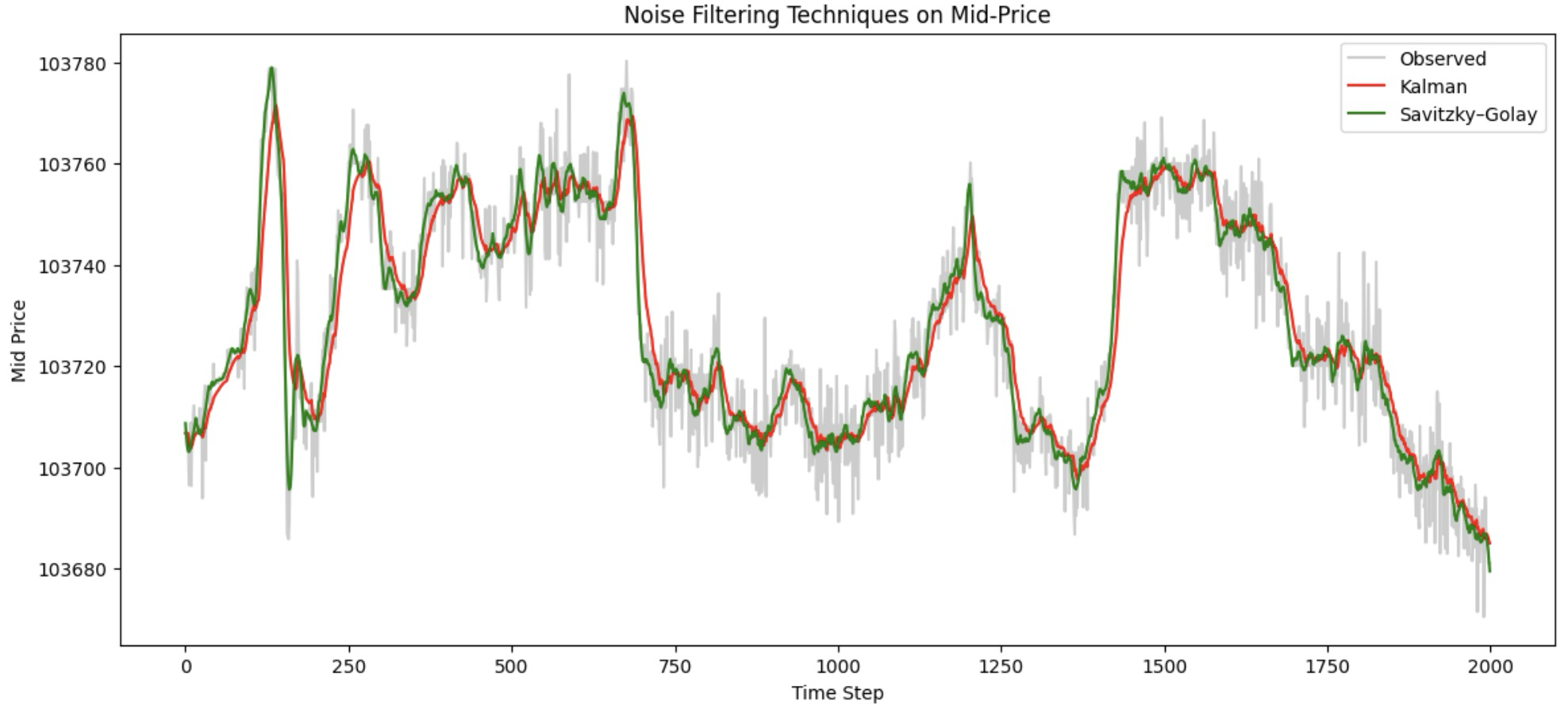}
    \caption{Savitzky–Golay and Kalman Filters on LOB}
    \label{fig:enter-label}
\end{figure}

% \medskip
\subsubsection{Savitzky–Golay Smoothing} 
Our first filter Savitzky–Golay fits local polynomials to sliding windows of data using least squares regression. And instead of averaging the observations, Savitzky–Golay aim to find a low degree polynomial. And by minimizing the local least‐squares error, this filter aim to preserve medium‐term trends while suppressing high‐frequency noise. In our code of execution, balancing between accuracy and latency of computation, we fit a cubic polynomial, dynamically in a rolling window. Specifically, at each time \(t\), fit a cubic polynomial\(d = 3\) over the window size \(2m+1 = 21\) by solving the following equation:

\begin{equation}
\min_{a_0, a_1, a_2, a_3} \; \sum_{j=-10}^{10} \left[v_{t+j} - \left(a_0 + a_1 j + a_2 j^2 + a_3 j^3 \right) \right]^2.
\end{equation}

The optimal coefficients \(\{a_\ell\}\) are obtained by solving the normal equations:
\begin{equation}
(\mathbf{A}^\top \mathbf{A})\,\mathbf{a} = \mathbf{A}^\top \mathbf{v},
\quad \text{where} \quad A_{j,\ell} = j^\ell,\quad j = -10, \dots, 10,\quad \ell = 0,\dots,3.
\end{equation}

Once the system is solved once for this fixed window, one can pre-compute convolution weights \( \{c_j\}_{j=-10}^{10} \) such that the smoothed estimate is:
\begin{equation}
\hat{v}_t = \sum_{j=-10}^{10} c_j\,v_{t+j}.
\end{equation}

\subsubsection{Kalman Filtering}
In comparison, we apply a one-dimensional Kalman filter to smooth each feature series \(v_t\), treating it as a noisy observation of a latent state \(x_t\). The model assumes a simple random walk for the latent state, where \(Q\) and \(R\) denote the variances of the process and observation noise.

\begin{equation}
  x_t = x_{t-1} + w_{t-1},\quad w_{t-1} \sim \mathcal{N}(0, Q),
  \qquad
  v_t = x_t + \varepsilon_t,\quad \varepsilon_t \sim \mathcal{N}(0, R),
\end{equation}

The Kalman filter recursively estimates the state \(x_t\) by updating the posterior mean and variance. The Kalman gain \(K_t\) is chosen to minimize the expected squared error of the estimate where \(\hat{x}_t\) is the smoothed estimate of the latent state at time \(t\), and \(P_t\) is the updated estimate variance.
\begin{equation}
  K_t = \frac{P_{t-1} + Q}{P_{t-1} + Q + R}, 
  \quad
  \hat{x}_t = \hat{x}_{t-1} + K_t \left(v_t - \hat{x}_{t-1} \right),
\end{equation}

\begin{equation}
  P_t = (1 - K_t)(P_{t-1} + Q),
\end{equation}

\bigskip

To assess the impact of denoising on model performance, we apply the same architecture and training pipeline across three versions of the dataset: raw, Savitzky–Golay smoothed, and Kalman filtered. By evaluating the classification accuracy under each setting, we aim to understand how well these filters enhance signal clarity and mitigate microstructural noise in high-frequency limit order book data.

\section{Results}
We use the \emph{F$_1$ score} to evaluate our model’s classification performance, and for each class $c$, it is defined as the harmonic mean of precision and recall:

\begin{equation}
  F_{1,c}
  = 2 \,\frac{\mathrm{Precision}_c \,\times\, \mathrm{Recall}_c}
           {\mathrm{Precision}_c + \mathrm{Recall}_c},
\end{equation}

\begin{align}
  \mathrm{Precision}_c &= \frac{\mathrm{TP}_c}{\mathrm{TP}_c + \mathrm{FP}_c}, 
  &
  \mathrm{Recall}_c    &= \frac{\mathrm{TP}_c}{\mathrm{TP}_c + \mathrm{FN}_c},
\end{align}

Where $\mathrm{TP}_c$, $\mathrm{FP}_c$, and $\mathrm{FN}_c$ denote the true positives, false positives, and false negatives for class $c$, respectively. And in our research, $\mathrm{TP}_c$ is the number of times the model correctly predicts class $c$ (e.g.\ an upward mid‐price change), $\mathrm{FP}_c$ is the number of times it predicts $c$ when the true movement is different, and $\mathrm{FN}_c$ is the number of times the true movement is $c$ but the model predicts otherwise. 

Here, $\mathrm{Precision}_c$ measures the fraction of predicted mid‐price moves of type $c$ that are correct, and $\mathrm{Recall}_c$ measures the fraction of actual moves of type $c$ that the model successfully detects.  Because increasing precision tends to reduce recall (and vice versa), the F$_1$ score provides a balanced metric by taking their harmonic mean.

And for the following result, we use the first 100\,000 LOB snapshots on 2025-01-30, with an 80\%/20\% train–test split (and 20\% of the training set for validation).  We discard any snapshot missing one of the 40 bid–ask levels, so for a certain testing set, we have a minimum of 5\,442 test observations. For example, in the following case, we see an overall accuracy of 0.715.

\bigskip

\subsection{Effect of Prediction Horizon on Three‐Class Prediction}

For the following table, we evaluate ternary (up/flat/down) classification performance across different filtering techniques and prediction horizons. We train our model on 80,000 rows and test on 20,000 rows of snapshots. Across each prediction interval and LOB depth, Savitzky filter generally helps to improve the prediction accuracy but there is no significant differences across models.

\begin{table}[H]
    \centering
    \begin{tabular}{lcccccc}
      \toprule
      Filter           & CatBoost & DeepLOB & CNN+LSTM & CNN+XGB & XGBoost & Logistic Reg. \\
      \midrule
      \multicolumn{7}{c}{\textbf{Next 100 ms, 5‐level LOB}} \\
      Raw              & 0.3922   & 0.4080   & 0.3987   & 0.4029  & 0.3939  & 0.3777        \\
      Kalman           & 0.3651   & 0.3838   & 0.3594   & 0.3609  & 0.3875  & 0.3732        \\
      Savitzky–Golay   & 0.4135   & \textbf{0.4281}   & 0.4207   & 0.4179  & 0.4173  & 0.4101        \\
      \midrule
      \multicolumn{7}{c}{\textbf{Next 500 ms, 40‐level LOB}} \\
      Raw              & 0.4066   & 0.4286   & 0.4160   & 0.4159  & 0.4457  & 0.4356        \\
      Kalman           & 0.3216   & 0.3044   & 0.3255   & 0.3211  & 0.4114  & 0.2993        \\
      Savitzky–Golay   & 0.5189   & 0.5237   & 0.5252   & 0.5282  & 0.5393  & \textbf{0.5434}        \\
      \midrule
      \multicolumn{7}{c}{\textbf{Next 1000 ms, 40‐level LOB}} \\
      Raw              & 0.4046   & 0.4099   & 0.4142   & 0.4118  & 0.4357  & 0.4901        \\
      Kalman           & 0.3191   & 0.3618   & 0.3603   & 0.3397  & 0.4084  & 0.3421        \\
      Savitzky–Golay   & 0.4611   & 0.4762   & 0.4936   & 0.4604  & 0.4999  & \textbf{0.5382}        \\
      \bottomrule
    \end{tabular}
    \caption{Ternary Classification Across Horizons and Filters (\(T=1\))}
    \label{tab:3class-summary}
  \end{table}
  
% \clearpage

% \FloatBarrier
% \clearpage

\subsection{Effect of Prediction Horizon on Two‐Class Prediction}
For the following table, we evaluate Binary (up/down) classification performance across different filtering techniques and prediction horizons. Similarly, we train our model on 80,000 rows and test on 20,000 rows of snapshots. Across each prediction interval and LOB depth, the Savitzky filter generally helps to improve the prediction accuracy, but there are no significant differences across models.

\begin{table}[H]
    \centering

\begin{tabular}{lcccccc}
\toprule
Filter            & CatBoost & DeepLOB & CNN+LSTM & CNN+XGB & XGBoost & Logistic Reg. \\
\midrule
\multicolumn{7}{c}{\textbf{Next 100 ms, 5‐level LOB}} \\
Raw               & 0.5257   & 0.5215  & 0.5235   & 0.5211  & 0.5296   & 0.5216        \\
Kalman            & 0.5129   & 0.5047  & 0.5043   & 0.5148  & 0.5268   & 0.5128        \\
Savitzky–Golay    & 0.5271   & 0.5296  & 0.5317   & 0.5327  & \textbf{0.5338}   & 0.5336        \\
\midrule
\multicolumn{7}{c}{\textbf{Next 500 ms, 40‐level LOB}} \\
Raw               & 0.5941   & 0.6333  & 0.6323   & 0.6320  & 0.6542   & 0.6517        \\
Kalman            & 0.5046   & 0.4825  & 0.5139   & 0.5716  & 0.6301   & 0.4882        \\
Savitzky–Golay    & 0.6260   & 0.7189  & 0.7189   & 0.7130  & 0.7281   & \textbf{0.7284}        \\
\midrule
\multicolumn{7}{c}{\textbf{Next 1000 ms, 40‐level LOB}} \\
Raw               & 0.5799   & 0.6448  & 0.6372   & 0.6276  & 0.6509   & 0.6515        \\
Kalman            & 0.5109   & 0.4908  & 0.4680   & 0.5536  & 0.6115   & 0.4930        \\
Savitzky–Golay    & 0.6140   & 0.6947  & 0.6888   & 0.6876  & \textbf{0.7150}   & 0.7089        \\
\bottomrule
\end{tabular}
    \caption{Binary Classification Across Horizons and Filters (\(T=1\))}
    \label{tab:3class-summary}
\end{table}

\subsection{Different Depths’ Impact and Data Coverage: XGBoost as Example}
\setlength{\tabcolsep}{4pt}
\renewcommand{\arraystretch}{0.9}

Given the results shown above, we proceeded with the settings that produced the highest predictive accuracy and continued exploring the effects of parameter tuning. Here we use XGBoost as an example for binary classification: we train our model on 80,000 snapshots and test it on 20,000 snapshots, all at 100ms resolution. All models use Savitzky–Golay–smoothed data, and we manually explore various LOB depths by dropping any snapshot that lacks bids and asks at the $k$th level, for $k$ from 40 down to 5 levels. The sample size increases from 5,442 to 18,336, but accuracy and F$_1$ scores decrease (0.715→0.580). This highlights a trade‐off between market coverage and predictive performance. Next steps include parameter tuning and robustness checks across depths.

\begin{table}[H]
  \centering
  \begin{tabular}{l c c c c}
    \toprule
    Depth      &  Accuracy &  F$_1$(0) &  F$_1$(1) &  Support \\
    \midrule
    40 levels  & 0.7150   & 0.7115   & 0.7184   & 5\,442   \\
    10 levels  & 0.5837   & 0.5732   & 0.5937   & 17\,471  \\
    5 levels   & 0.5797   & 0.5560   & 0.6009   & 18\,336  \\
    \bottomrule
  \end{tabular}
  \caption{Prediction performance versus LOB depth (SG filter, $T=1$). Shallower LOB increases coverage but reduces accuracy and F$_1$.}
  \label{tab:depth-results}
\end{table}

\subsection{Model Comparison: Sequence Length Effects: XGBoost and Logistic Regregression }
  \setlength{\tabcolsep}{4pt}
  \renewcommand{\arraystretch}{0.9}
    Here, \(T\) is the number of sequential LOB snapshots fed into the model to forecast the next mid‑price move. Note that our data use a 100ms resolution: when \(T =1\), the model uses a single 100ms snapshot to predict the mid‑price change over the next 1000ms; when \(T = 10\), the model is trained on a concatenated vector of ten 100ms snapshots (spanning 1s), enabling it to exploit temporal patterns to predict the mid‑price change one second after the sequence start.
    
    We feed each model with 1 million snapshots (5 bid levels and 5 ask levels) at 100ms resolution, smoothed with a Savitzky–Golay filter. We train on 800,000 rows and test on 200,000 rows; however, only 148,340 rows (73,953 + 74,387, i.e.\ 75\(\%\) coverage) contain at least five bids and five asks. The result highlights a trade‑off between predictive power and computational efficiency: training on sequences instead of single snapshots increases F1 scores for both models, yielding an approximately 2\(\%\) accuracy gain.

  \begin{table}[H]
    \centering
    % \scriptsize                % make font smaller
    \begin{tabular}{l c c c c c c c}
      \toprule
      Model             & T  & Accuracy & F$_1$(0) & F$_1$(1) & Support 0 & Support 1 & Run time     \\
      \midrule
      XGBoost           & 1  & 0.5732   & 0.5757   & 0.5707   & 73\,953   & 74\,387   & 1 m 36 s     \\
      XGBoost           & 10 & 0.5949   & 0.5989   & 0.5909   & 73\,951   & 74\,387   & 7 m 04 s     \\
      Logistic Reg.     & 1  & 0.5551   & 0.5442   & 0.5656   & 73\,953   & 74\,387   & 1 m 11 s     \\
      Logistic Reg.     & 10 & 0.5716   & 0.5654   & 0.5777   & 73\,951   & 74\,387   & 9 m 13 s     \\
      \bottomrule
    \end{tabular}
    \caption{Comparison of model performance and runtimes (SG filter) across sequence lengths (Run time: time to train the model on 1 million data points on M2 chip MacBook}
    \label{tab:model-comparison}
  \end{table}

\section{Conclusion}
The results above offer an alternative perspective to much of the existing literature as adding an extra CNN layer to an CNN + LSTM based architecture did not yield the substantial accuracy gains we expected; rather, data quality, noise‐handling assumptions, and training methodology drive performance, allowing simpler models like XGBoost and Logistic Regression to match complex architectures with faster latency on high‐frequency, snapshot‐level LOB data.

Across both ternary and binary classification (see Subsections 5.1 and 5.2), all models consistently perform best after Savitzky–Golay smoothing. Note that the Kalman filter results are sometimes worse than using the raw data itself. And this is because the Kalman filter requires more extensive parameter tuning, which, in our project, we addressed through a limited grid search over a smaller sample and then fixed the parameters for concerns on computation limits. So unlike the Savitzky–Golay filter, whose smoothing window can be adjusted more dynamically, the Kalman filter’s parameters—such as process and measurement noise covariances—are more rigid and sensitive. As a result, the Kalman filter may not be the most suitable choice in our case. Furthermore, binary and ternary prediction accuracy ranges from 0.42 to 0.71—spanning 100ms predictions with 5‑level LOB data to 1000ms predictions with 40‑level LOB data.

And we found that under feature engineering and denoising, simpler models such as XGBoost and logistic regression marginally outperform our more complex neural networks by 1–2\%. And for practical concerns, in our research, we didn't experiment with deeper CNN architectures (more than three Conv layers or more than 128 dimensions each layer) since the increased training and inference time would likely erase any alpha(prediction edge). For instance, training a fancy network model that achieves 80\% prediction accuracy but takes 2 seconds to generate a forecast for the next 1 second, by the time it outputs the prediction, the opportunity has already faded. Thus, blindly chasing higher accuracy without accounting for model complexity and latency can be wasted.

Interestingly, for simpler benchmark models, data preprocessing and parameter choices (e.g., $T=1$ vs.\ $T=10$, prediction horizon, LOB depth) have a greater impact on performance than model complexity. These models train in a fraction of the time needed by the neural networks, making them more practical for real‑time use. Because our experiments were conducted on a single trading day (\texttt{2025-01-30}) over roughly 100,000 snapshots (\(\approx\) 166min) for Subsections 5.1–5.3 and 1,000,000 snapshots (\(\approx\) 28h) for Subsection 5.4, further testing on additional days and under different market conditions is needed to assess robustness.

All experiments were performed offline in Python; real‑time deployment will require further evaluation and fine tuning, and developing a trading strategy with the price prediction edge, and hardware acceleration. Future work should explore adaptive denoising techniques,explore the differences and similarities between working with transaction-level data versus snapshot-based datasets, and extend analysis to less liquid cryptocurrency and other assets' limit order books, where supply–demand imbalance signals may exhibit different dynamics.

\FloatBarrier
\section{Acknowledgment}
I would like to express my sincere gratitude to my advisor, Professor Anthony Lee Zhang, Kotaro Yoshida, and Victor Lima for their invaluable guidance and support throughout this research. 

\FloatBarrier

\section{References}

% \begin{thebibliography}{99}

\end{document}